\documentclass[a4paper, 10pt, conference]{ieeeconf}      
\usepackage{FG2020}

\usepackage{graphicx}
\usepackage{amssymb,multirow,xcolor,textcomp,makecell}
\usepackage{algorithm}
\usepackage{dblfloatfix}

\usepackage{algorithm}
\usepackage{amsmath}
\usepackage{subfig}
\usepackage{multirow}
\usepackage{color}
\usepackage{lscape,adjustbox}
\usepackage{slashbox}
\usepackage{float}
\usepackage{amssymb}
\usepackage{tabu}

\usepackage{lineno}
\usepackage{algorithm}
\usepackage{amsmath}
\usepackage{subfig}
\usepackage{multirow}
\usepackage{color}
\usepackage{lscape,adjustbox}
\usepackage{float}
\usepackage{amssymb}

\usepackage{xcolor}
\usepackage{amsmath}
\usepackage[noend]{algpseudocode}

\usepackage{environ}

\NewEnviron{myequation}{%
\begin{equation}
\scalebox{0.75}{$\BODY$}
\end{equation}
}

\NewEnviron{myequation2}{%
\begin{equation}
\scalebox{0.95}{$\BODY$}
\end{equation}
}

\FGfinalcopy 

\def\FGPaperID{****} 

\title{\LARGE \bf
Multi-view Deep Features for Robust Facial Kinship Verification 
}

\author{\parbox{16cm}{\centering
    {\large Oualid Laiadi$^{1,4}$, Abdelmalik Ouamane$^2$, Abdelhamid Benakcha$^3$, \\Abdelmalik Taleb-Ahmed$^4$ and Abdenour Hadid$^5$}\\
    {\normalsize
    $^1$ Laboratory of LESIA, University of Biskra, Algeria\\
    $^2$ Laboratory of LI3C, University of Biskra, Algeria\\
		$^3$ Laboratory of LGEB, University of Biskra, Algeria\\
		$^4$ Univ. Polytechnique Hauts-de-France, CNRS, Univ. Lille, ISEN, Centrale Lille, UMR 8520 - IEMN - DOAE, F-59313 Valenciennes, France.\\
		$^5$ Center for Machine Vision and Signal Analysis, University of Oulu, Finland}}
}

\begin{document}

\ifFGfinal
\thispagestyle{empty}
\pagestyle{empty}
\else
\author{Anonymous FG 2020 submission\\ Paper ID \FGPaperID \\}
\pagestyle{plain}
\fi
\maketitle

\begin{abstract}

Automatic kinship verification from facial images is an emerging research topic in machine learning community. In this paper, we proposed an effective facial features extraction model based on multi-view deep features. Thus, we used four pre-trained deep learning models using eight features layers (FC6 and FC7 layers of each VGG-F, VGG-M, \linebreak VGG-S and VGG-Face models) to train the proposed Multilinear Side-Information based Discriminant Analysis integrating Within Class Covariance Normalization (MSIDA+WCCN) method. Furthermore, we show that how can metric learning methods based on WCCN method integration improves the Simple Scoring Cosine similarity (SSC) method. We refer that we used the SSC method in RFIW'20 competition using the eight deep features concatenation. Thus, the integration of WCCN in the metric learning methods decreases the intra-class variations effect introduced by the deep features weights. We evaluate our proposed method on two kinship benchmarks namely KinFaceW-I and KinFaceW-II databases using four Parent-Child relations (Father-Son, Father-Daughter, \linebreak Mother-Son and Mother-Daughter). Thus, the proposed MSIDA+WCCN method improves the SSC method with 12.80\% and 14.65\% on KinFaceW-I and KinFaceW-II databases, respectively. The results obtained are positively compared with some modern methods, including those that rely on deep learning.  


\end{abstract}

\section{Introduction}
The basic idea of automatic kinship verification using facial images is to check if a given two facial images input have pertinence from the same family or not. Several applications can be useful under automatic kinship verification e.g. for forensics, finding missing children, social media comprehension and image annotation. Thus, a DNA test is the most reliable source for kinship verification, it unfortunately cannot be utilized in many situations such as in video surveillance.

Many authors feed their method by different features or multiple features (multi-view data) to represent facial images for kinship verification. Lu et al. used the Multiview neighborhood repulsed metric learning (MNRML) ~\cite{r9.Lu:2014:NRM:2574225.2574494} method to train four multi-view features, Local Binary Patterns (LBP), Learning-based descriptor (LE), SIFT and Three-patch LBP (TPLBP). Yan et al. ~\cite{r10.6824230} employed three different feature descriptors including Local Binary Patterns(LBP), Spatial Pyramid LEarning (SPLE) and Scale-Invariant Feature Transform (SIFT) to extract different and complementary information from each face image through DMML method. Yan et al. ~\cite{r120.6981937} applied three dif-ferent feature  descriptors including LBP, spatial pyramid lEarning (SPLE), and SIFT to extract different and complementary information from each face image to train the MPDFL method. Lu et al. ~\cite{80.7953665} used four features as it; Local Binary Patterns (LBP), Dense SIFT (DSIFT), the histogram of oriented gradients (HOG) and LPQ for train DDMML method. Lu et al. ~\cite{r106.8027090} used MvDML to train four multi-view features, Local Binary Patterns (LBP), Learning-based descriptor (LE), SIFT and Three-patch LBP (TPLBP). Laiadi et al. ~\cite{Laiadi2019} used three features LPQ, BSIF and CoALBP to train SIEDA method. Dornaika et al. used MNRML to train the two features, FC7 layers of VGG-F and VGG-Face for the purpose of kinship verification. Laiadi et al. proposed TXQDA ~\cite{LAIADI2019_3} method to train LPQ and BSIF features using ten scales.

In this work, we propose a new framework to kinship verification from facial images using eight deep features based on four pre-trained deep learning networks. For this reason, we extract FC6 and FC7 layers from VGG-F, VGG-M, VGG-S and VGG-Face models to train the proposed Multilinear Side-Information based Discriminant Analysis integrating Within Class Covariance Normalization (MSIDA+WCCN) method. We report our preliminary experimental investigations on the KinFaceW-I and KinFaceW-II benchmarks using four relations, Father-Son, Father-Daughter, Mother-Son and Mother-Daughter face subsets showing very high performance compared to state-of-the-art methods.

\section{Proposed Framework}

\begin{figure}[H]
\captionsetup{font=scriptsize}
\centering
\includegraphics[width=\linewidth]{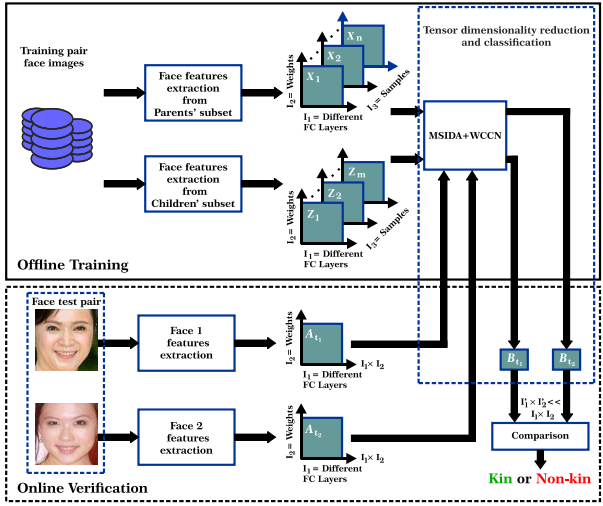}
\caption{Block diagram of the proposed face pair matching system.}
\label{system}
\end{figure}

Figure \ref{system} depicts an overview of our proposed framework. The input is a pair of two face images e.g. a Parent and a Child. We extract features from these images into eight deep features of the input test pair. We compute the cosine similarity between the two facial images to encode the final metric. The cosine similarity is fed to the ROC curve for the performance evaluation. 

\subsection{Extracting Multi-view Deep Features}

\begin{figure}[!htb]
\captionsetup{font=scriptsize}
\centering
\includegraphics[width=\linewidth]{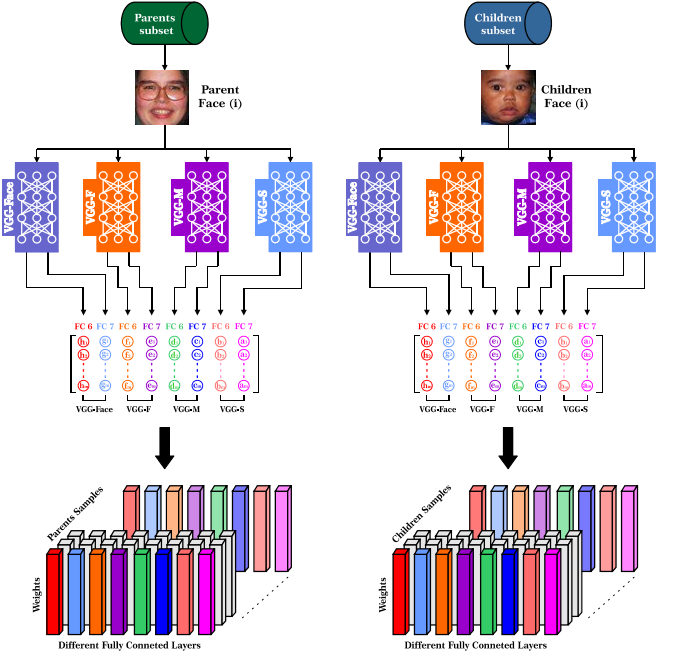}
\caption{Multi-view deep features extraction (Mv-VGG) and tensor design.}
\label{fig:1}
\end{figure}

Many methods suggested in the literature on automatic verification of kinship have focused mainly on analyzing deep features trained on facial images (i.e. VGG-Face), thus ignoring deep features trained on object images (i.e. VGG-F, VGG-M and VGG-S). Recently, deep facial features have shown great performance than their shallow counterparts to verify kinship relation (e.g. ~\cite{ZHOU201984}). When considering the facial deep information, the problem usually consists in learning a discriminating metric where the classification (e.g. kinship verification in our case) becomes more affordable when combined to the object deep features. As suggested in ~\cite{Dornaika2019}, we consider the object deep features for kinship verification. Facial and object features show a complementarity which extracted by MSIDA+WCCN method. This is in contrast to use facial deep features or object deep features information separately. Therefore, we extract the facial deep features using VGG-Face ~\cite{r3} method and object deep features \cite{DBLP:journals/corr/ChatfieldSVZ14} (VGG-F, VGG-M and VGG-S methods) using MSIDA+WCCN method. Figure \ref{fig:1} depicts multi-view deep features extraction (Mv-VGG) and tensor design. From this figure, the different colors of block architecture represent the difference in each architecture type. For deep object features, VGG-Fast is an architecture contains number of parameters smaller than VGG-Medium, and this latter is an architecture contains number of parameters smaller than VGG-Slow and these three architectures were trained on the object recognition ILSVRC-2012 database. For deep face features, the VGG-Face architecture was trained on VGG Face database [17] which contains 2.6M facial images from 2,622 identities. Furthermore, in the tensor representation, the length of each data stacked in a tensor mode must be the same, and this property was saved by the four pre-trained models with 4096 neurons in each of the eight used fully connected layers (i.e the FC6 and FC7 layers of the four pre-trained models have the same length).

\section{Multilinear Side-Information based Discriminant Analysis integrating Within Class Covariance Normalization (MSIDA+WCCN)}\label{relWork}
\subsection{Side-Information based Linear Discriminant analysis (SILD)}
The positive classes pair images are directly utilized to calculate the within class scatter matrix and the negative classes pair images are used to compute the between class scatter matrix. Let us refer that $P_{class} =\left\{(\check{\xi}^1_i, \hat{\xi}^1_i) : l(\check{\xi}^1_i) = l(\hat{\xi}^1_i)\right\}$ as the collection of positive-class image pairs and \\ $N_{class} =\left\{(\check{\xi}^0_i, \hat{\xi}^0_i) : l(\check{\xi}^0_i) 
\neq l(\hat{\xi}^0_i)\right\}$ as the collection of negative-class image pairs, where the image $\xi$ is represented by the class label $l(\xi)$. Here, the within-class and between-class scatter matrices of Side-Information based Linear Discriminant analysis ~\cite{r29.BMVC.25.125:abbreviated} (SILD) method can be represented by:

\begin{equation} \label{eq_9}
S_w^{sild}=\sum_{i=1}^{C_1}(\check{\xi}_{i}^{1}-\hat{\xi}_{i}^{1})(\check{\xi}_{i}^{1}-\hat{\xi}_{i}^{1})^T
\end{equation} 

\begin{equation} \label{eq_10}
S_b^{sild}=\sum_{i=1}^{C_0}(\check{\xi}_{i}^{0}-\hat{\xi}_{i}^{0})(\check{\xi}_{i}^{0}-\hat{\xi}_{i}^{0})^T
\end{equation}  

The target function for SILD is:

\begin{equation} \label{eq_11}
\begin{array}{c}
W_{opt}^{sild}=argmax_{W}\frac{W^TS^{sild}_bW}{W^TS^{sild}_wW}\\
\end{array}
\end{equation}

The problem in \ref{eq_11} can be solved by a two-step method ~\cite{r112.531802}. Firstly, $S_w$ is diagonalized as follows:

\begin{equation} \label{eq_32}
S_w = H\Lambda H^T
\end{equation}

\begin{equation} \label{eq_33}
(H\Lambda^{-1/2})^T S_w (H\Lambda^{-1/2}) = I
\end{equation}

Secondly, $S_b$ is also diagonalized:

\begin{equation} \label{eq_34}
(H\Lambda^{-1/2})^T S_b (H\Lambda^{-1/2}) = ZEZ^T
\end{equation}

Finally, the projection matrix can be computed as:

\begin{equation} \label{eq_35}
W^{sild} = H\Lambda^{-1/2}Z
\end{equation}
where H and Z are orthogonal matrices and $\Lambda$ and E are diagonal matrices.

A solution to the optimization problem in (\ref{eq_11}) is obtained via solving the generalized eigenvalue problem. The projection matrix of SILD is formed by the first $v$ eigenvectors in (\ref{eq_35}) that ordered in the descending order of eigenvalues.

\subsection{Proposed Multilinear Side-Information based Discriminant Analysis integrating Within Class Covariance Normalization}
Let a Tensor training set $\{\textbf{\textit{X}}, \textbf{\textit{Z}}\}$ of $\mathrm{c}$ classes, where: ${\textbf{\textit{X}}}\in\mathrm{\Re^{I_1\times I_2\times\cdots \times I_N \times c}}$ contains $\mathrm{c}$ samples of Parents samples and ${\textbf{\textit{Z}}}\in\mathrm{\Re^{\mathrm{I_1}\times \mathrm{I_2}\times\cdots \times \mathrm{I_N} \times \mathrm{c}}}$ contains $\mathrm{c}$ samples of Children samples. The goal of MSIDA ~\cite{r1} is the calculation of $\mathrm{N}$ projection matrices ($W_1  \in \Re^{\mathrm{I_1} \times \mathrm{I^{'}_{1}}}, W_2 \in \Re^{\mathrm{I_2} \times \mathrm{I^{'}_{2}}}, \ldots, W_\mathrm{N} \in \Re^{\mathrm{I_N} \times \mathrm{I^{'}_{N}}}$). Thus, we calculate one projection matrix for each tensor mode. 
The objective function of MSIDA method is defined as follow:


\begin{equation} \label{eq_44} 
J(W_\mathrm{k}) = \frac{W_{\mathrm{k}}^{T}S_b^\mathrm{k} W_\mathrm{k}}{W_{\mathrm{k}}^{T}S_w^\mathrm{k} W_\mathrm{k}} 
\end{equation}

We calculate the two covariance matrices $S_{b}^\mathrm{k}$ and $S_{w}^\mathrm{k}$ for each $\mathrm{k}$ mode by:

\begin{myequation} \label{eq_45}  
S_{w} = \sum\limits_{\mathrm{p=1}}^{\prod_{\mathrm{o\neq k}}\mathrm{I_{o}}} S_{w}^{\mathrm{p}}, S_{w}^{\mathrm{p}} = \sum_{i=1}^{C_1}((\check{\xi}_{i}^{1})^{k,p}-(\hat{\xi}_{i}^{1})^{k,p})((\check{\xi}_{i}^{1})^{k,p}-(\hat{\xi}_{i}^{1})^{k,p})^T
\end{myequation}

\begin{myequation} \label{eq_46}  
S_{b} = \sum\limits_{\mathrm{p=1}}^{\prod_{\mathrm{o\neq k}}\mathrm{I_{o}}} S_{b}^{\mathrm{p}}, S_{b}^{\mathrm{p}} = \sum_{i=1}^{C_0}((\check{\xi}_{i}^{0})^{k,p}-(\hat{\xi}_{i}^{0})^{k,p})((\check{\xi}_{i}^{0})^{k,p}-(\hat{\xi}_{i}^{0})^{k,p})^T
\end{myequation}

Now that the solution for one mode is known, the optimization problem in equation \ref{eq_44} can be solved iteratively. The projection matrices $W_1, W_2, \ldots, W_\mathrm{N}$ are first initialized to identity. At each iteration $W_1, W_2, \ldots, W_{\mathrm{k-1}}, W_{\mathrm{k+1}}, \ldots W_\mathrm{N}$ are hypothetical known and $W_\mathrm{k}$ is estimated. Set: $\textbf{\textit{U}} = \textbf{\textit{X}} \times_1 W_1 \ldots \times_{\mathrm{k-1}} W_{\mathrm{k-1}}\times_{\mathrm{k+1}} W_{\mathrm{k+1}} \ldots \times_{\mathrm{N}} W_\mathrm{N}$ and $\textbf{\textit{Y}} = \textbf{\textit{Z}} \times_1 W_1 \ldots \times_{\mathrm{k-1}} W_{\mathrm{k-1}}\times_{\mathrm{k+1}} W_{\mathrm{k+1}} \ldots \times_{\mathrm{N}} W_\mathrm{N}$  are replaced in equation \ref{eq_44} by $\textbf{\textit{X}}$ and $\textbf{\textit{Z}}$. The new equation can be solved by the generalized eigenvalue decomposition problem:

\begin{equation} \label{eq_47} 
S_{b}^\mathrm{k} W_\mathrm{k} = \Lambda_\mathrm{k} S_{w}^\mathrm{k} W_\mathrm{k}
\end{equation}

Where, $W_\mathrm{k}$ is the eigenvectors matrix and $\Lambda_\mathrm{k}$ the eigenvalues matrix.

The iterative process of MSIDA breaks up on the recognition of one of the following situations: i) The number of iterations reaches a predefined maximum; or ii) the difference of the estimated projection between two consecutive iterations is less than a threshold, $\left\|W_\mathrm{k}^{\mathrm{iteration}}-W_\mathrm{k}^{\mathrm{iteration-1}}\right\|< \mathrm{I_k I_{k}\mathrm{\epsilon}}$ where $\mathrm{I_k}$ is the $\mathrm{k}$ mode dimension of $W_{\mathrm{k}}^{\mathrm{iteration}}$. As depicted in Fig. \ref{system}, the block diagram of the proposed approach consists of three essential components: feature extraction, tensor subspace transformation and comparison. We focus in this work on subspace transformation and the feature extraction based multiple scales local descriptor. 

\subsection{Within-Class Covariance Normalization}
The first use of the Within-Class Covariance Normalization (WCCN) is in the community of speaker recognition. While Dehak \textit{et al.} ~\cite{r75.5545402} founded that it is the best technique to project the reduced-vectors of LDA method to a new subspace determined by the square-root of the inverse of the within-class covariance matrix. We propose a new variant of MSIDA by integrating WCCN:

\begin{myequation2} \label{eq_18}
G = \sum\limits_{\mathrm{p=1}}^{\prod_{\mathrm{o\neq k}}\mathrm{I_{o}}} G^{\mathrm{p}}, G^{\mathrm{p}} =\sum_{i=1}^{C_1} \frac{(W^{k})^T(\check{\xi}_{i}^{1})^{k,p}-(W^{k})^T(\hat{\xi}_{i}^{1})^{k,p}}{(W^{k})^T(\check{\xi}_{i}^{1})^{k,p}-(W^{k})^T(\hat{\xi}_{i}^{1})^{k,p}}
\end{myequation2} 

where, $W^{k}$ is the MSIDA projection matrix found in Eq.\ref{eq_47}. The WCCN projection matrix $C$ is obtained by Cholesky decomposition ~\cite{r113.doi:10.1080/03610918208812265,r114.4808223} of the inverse of $G^k$: $(G^k)^{-1} = C^k (C^k)^T$. Where the new projection matrix $D^{k}$ is obtained by: $D^{k} = (C^k)^T W^{k}$. By imposing upper bounds on the classification error metric ~\cite{r110.6751354}, WCCN decreases the within-class variations effect by reducing the expected classification error on the training step.


\section{Experimental Analysis}

For experimental evaluation, we considered the KinFaceW-I and KinFaceW-II databases are gathered through Internet research, including some public figures with their parents and/or children. In the KinFaceWI dataset, there are 156, 134, 116, and 127 pairs corresponding to the F-S, F-D, M-S, and M-D relations, respectively. For the KinFaceW-II dataset, each kin relation type contains 250 pairs. In total KinFaceW-I count 1066 face images and 2000 face images for KinFaceW-II.

\begin{table*}[htbp]
\captionsetup{font=scriptsize}
  \centering
  \caption{Performance comparisons (\%) with state-of-the-art methods on KinFaceW-I and KinFaceW-II databases.}
    \begin{tabular}{c||c c c c c||c c c c c}
    \hline
    \multicolumn{1}{c||}{\multirow{2}[4]{*}{Method}} & \multicolumn{5}{c||}{KinFaceW-I} & \multicolumn{5}{c}{KinFaceW-II} \\
\cline{2-11}          & \multicolumn{1}{c}{F-S} & \multicolumn{1}{c}{F-D} & \multicolumn{1}{c}{M-S} & \multicolumn{1}{c}{M-D} & \multicolumn{1}{c||}{Mean} & \multicolumn{1}{c}{F-S} & \multicolumn{1}{c}{F-D} & \multicolumn{1}{c}{M-S} & \multicolumn{1}{c}{M-D} & \multicolumn{1}{c}{Mean} \\
    \hline
		
		MNRML ~\cite{r9.Lu:2014:NRM:2574225.2574494} &  72.50  &   66.50  &  66.20  &   72.00 & 69.90 & 76.90 & 74.30 & 77.40 & 77.60 & 76.50 \\
		DMML ~\cite{r10.6824230} &   74.50    &    69.50   &   69.50    &   75.50    &   72.25    &     78.50 & 76.50 & 78.50 & 79.50 & 78.25 \\
		MPDFL ~\cite{r120.6981937} &   73.50    &   67.50    &   66.10    &   73.10    &   70.10    &   77.30 & 74.70 & 77.80 & 78.00 & 77.00 \\ 
	  MMTL ~\cite{QIN2016350} &   N.A    &  N.A     &  N.A     &    N.A   &  73.70     &  N.A & N.A & N.A & N.A & 77.20 \\
		DDMML ~\cite{80.7953665} &  \textbf{86.40}     &   79.10    &   81.40    &   87.00    &  83.50 & 87.40 & 83.80 & 83.20 & 83.00 & 84.30 \\ 
		NRCML ~\cite{YAN201791} &   66.10    &   61.10    &   66.90    &    73.00   &   66.30    &     79.80 & 76.10 & 79.80 & 80.00 & 78.70 \\ 
		MKSM ~\cite{r105.ZHAO2018247} &  83.65     &   81.35    &   79.69    &   81.16    &   81.46    &  83.80 & 81.20 & 82.40 & 82.40 & 82.45 \\ 
		MvDML ~\cite{r106.8027090} &   /    &   /    &   /    &    /   &    /   &   80.40 & 79.80 & 78.80 & 81.80 & 80.20 \\
		Deep+Shallow ~\cite{BordalloLopez2018} &  68.80    &   68.80  &  70.50    &   65.50   &  68.40   & 66.50  & 68.80 & 65.40 & 65.40 & 66.50 \\
		$\mathrm{L^2M^3L}$ ~\cite{7894195} &   /    &   /    &   /    &    /   &   /    &     82.40 & 78.20 & 78.80 & 80.40 & 80.00 \\
		ResNet + CF ~\cite{8337841} &   78.00     &   83.70  &   87.00  & 80.80   & 82.40   &  87.70 & \textbf{86.00} & 86.70 & 87.40 & 86.60 \\
		RDML ~\cite{8522031} &  76.20& 74.20& 76.90& 82.20& 77.30&  79.30& 72.30& 77.40& 78.30& 76.80 \\
		
		MNRML+SVM \cite{Dornaika2019} & 85.90 & 79.85 & 86.20 & 86.62 & 84.55 & 87.20 & 82.60 & \textbf{88.40} & \textbf{89.40} & 86.90 \\
		SILD+WCCN/LR ~\cite{Laiadi2019_2} &   /    &   /    &   /    &    /   &   /    &     88.40 & 84.20 & 85.80 & 86.40 & 86.20 \\
		KML ~\cite{ZHOU201984} &   N.A    &  N.A     &  N.A     &    N.A   &  82.80     &  N.A & N.A & N.A & N.A & 85.70 \\
		MvGMML ~\cite{8803754} &    69.25 & 73.12 & 69.40 & 72.76 & 71.13 &     70.40 & 73.40 & 65.80 & 69.20 & 69.70 \\
    \hline
		SSC	&71.57	&70.83	&77.12	&79.88	&74.85  &72.80	&69.00	&73.80	&73.80	&72.35 \\
SILD	&73.75	&71.25	&76.25	&77.49	&74.69   &72.80	&69.20	&74.00	&74.00	&72.50 \\
MSIDA	&73.00	&72.96	&78.41	&77.91	&75.57		&75.00	&69.40	&75.80	&74.40	&73.95 \\
SILD+WCCN	&75.72	&72.39	&79.80	&80.74	&77.16   &77.40	&75.60	&75.80	&78.40	&76.80 \\

		\hline
\textbf{MSIDA+WCCN}	&85.98	&\textbf{85.93}	&\textbf{90.05}	&\textbf{88.62}	&\textbf{87.65}		&\textbf{89.40}	&82.80	&87.80	&88.00	&\textbf{87.00} \\
    \hline
    \end{tabular}%
  \label{tab:1}%
\end{table*}%


\begin{figure*}[!b]
\captionsetup{font=scriptsize}
   \centering
   \subfloat[][]{\includegraphics[width=.23\textwidth]{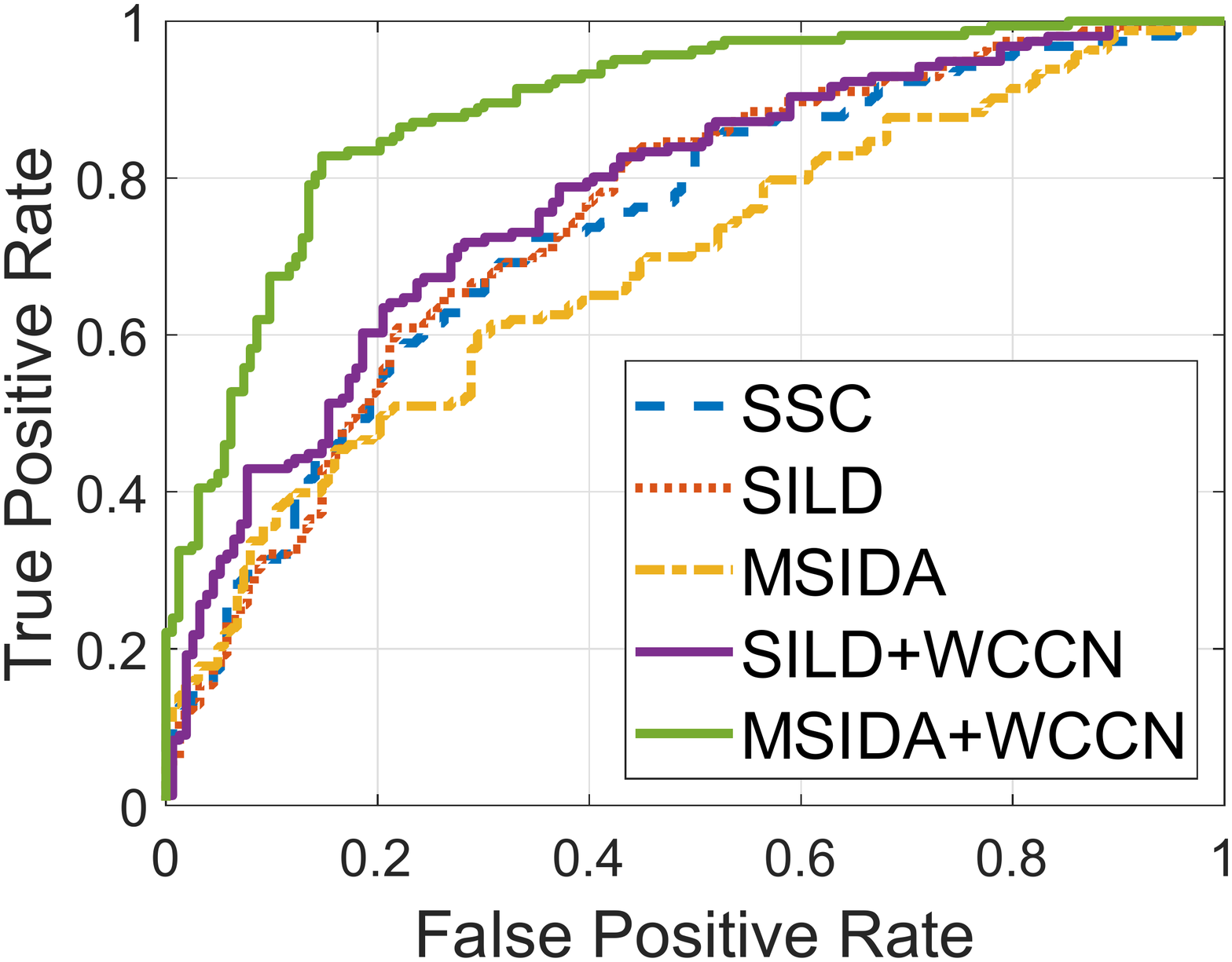}}\quad
   \subfloat[][]{\includegraphics[width=.23\textwidth]{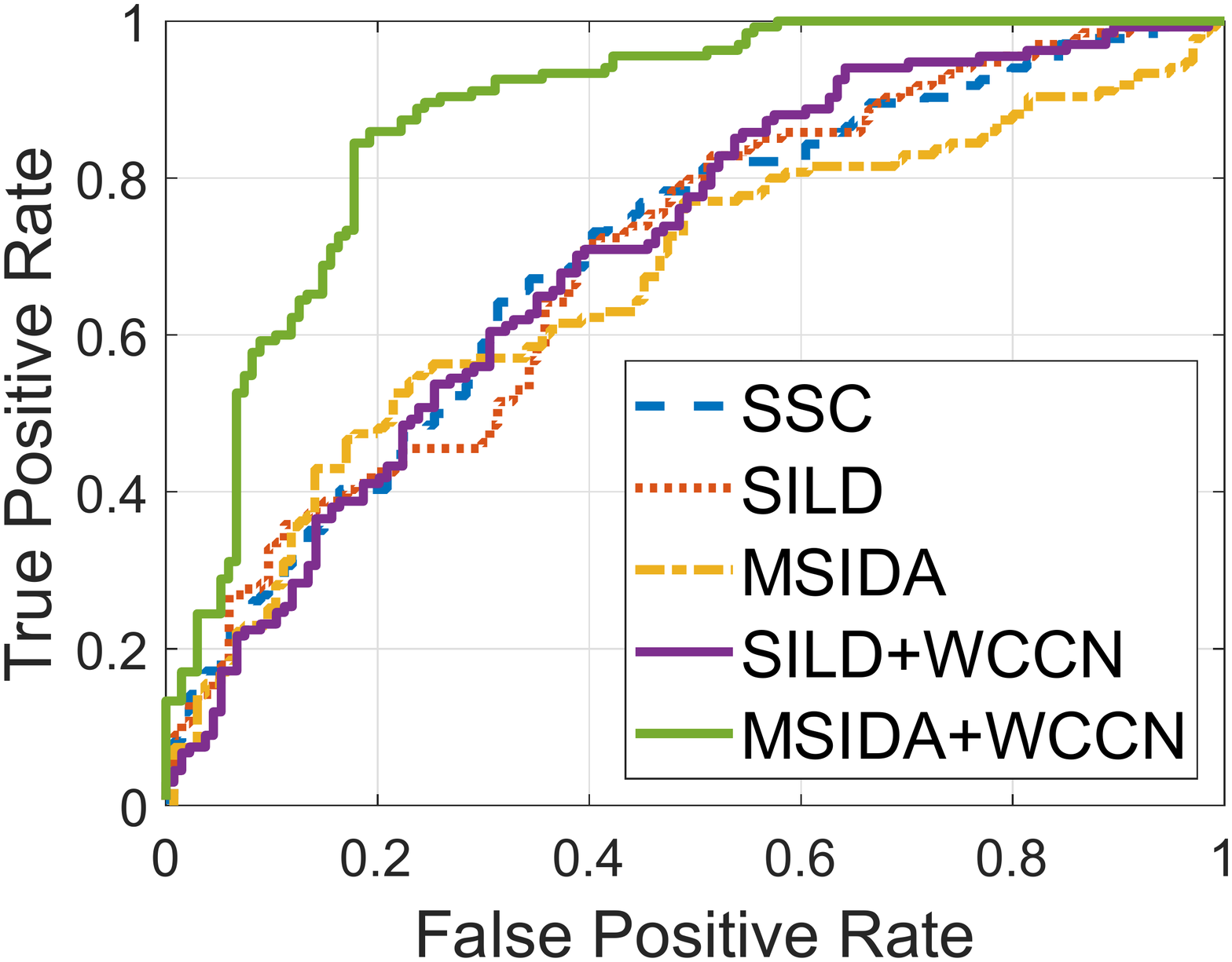}}\quad
   \subfloat[][]{\includegraphics[width=.23\textwidth]{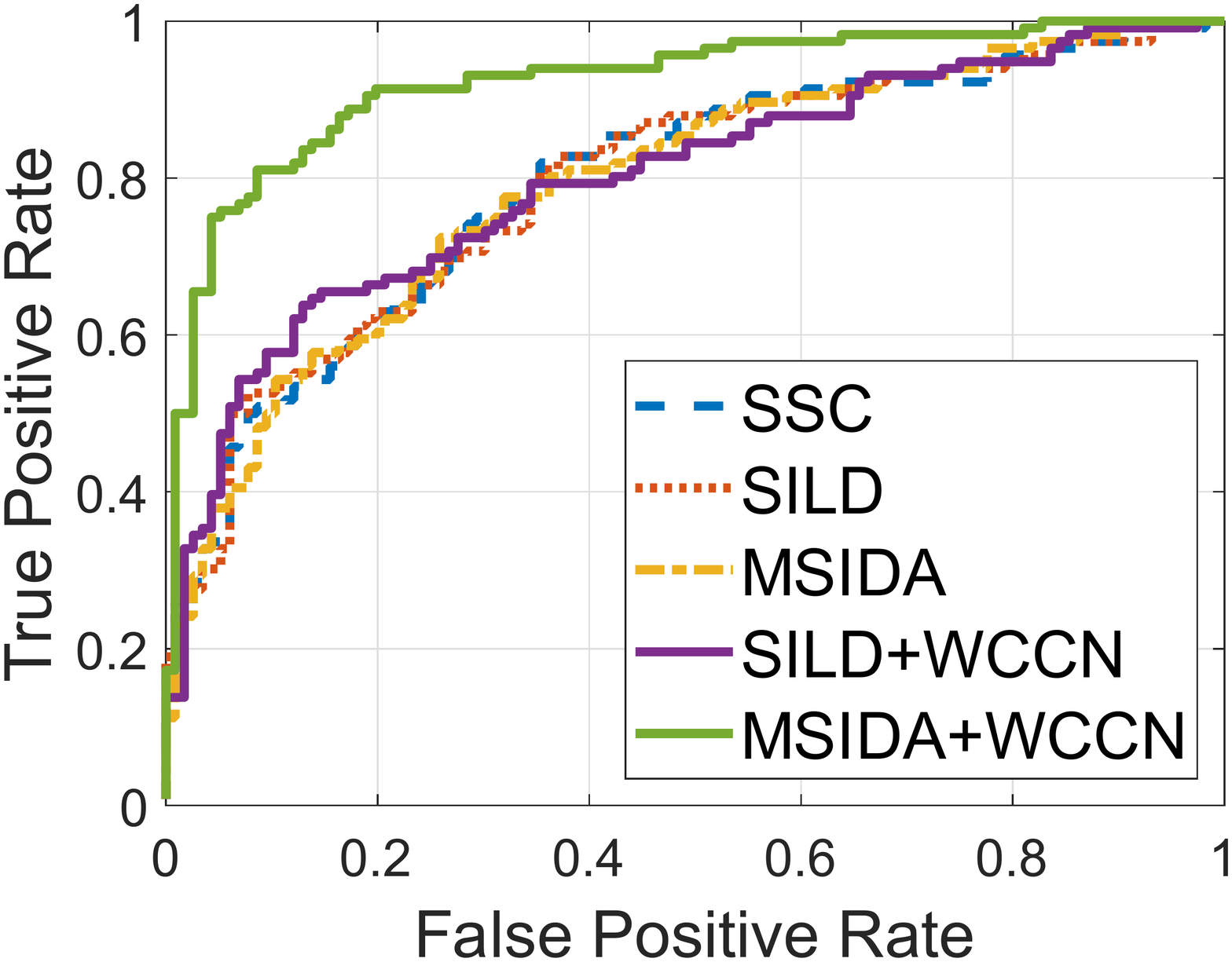}}\quad
   \subfloat[][]{\includegraphics[width=.23\textwidth]{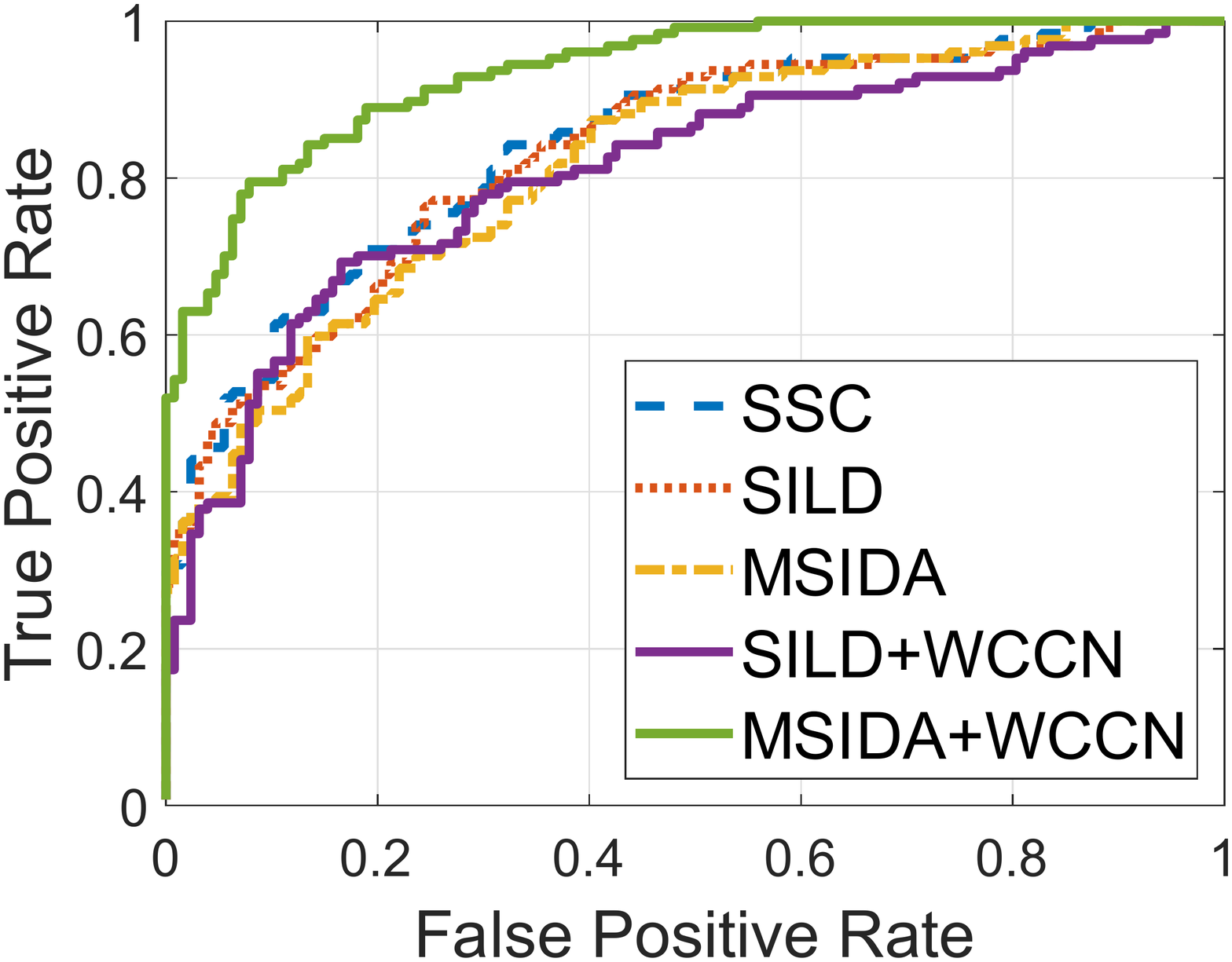}}
   \caption{ROC curves of different methods (SSC, SILD, MSIDA, SILD+WCCN and MSIDA+WCCN) on KinFaceW-I database obtained on (a) F-S set, (b) F-D set, (c) M-S set and (d) M-D set.}
   \label{fig:2}
\end{figure*}

\begin{figure*}[!b]
\captionsetup{font=scriptsize}
   \centering
   \subfloat[][]{\includegraphics[width=.23\textwidth]{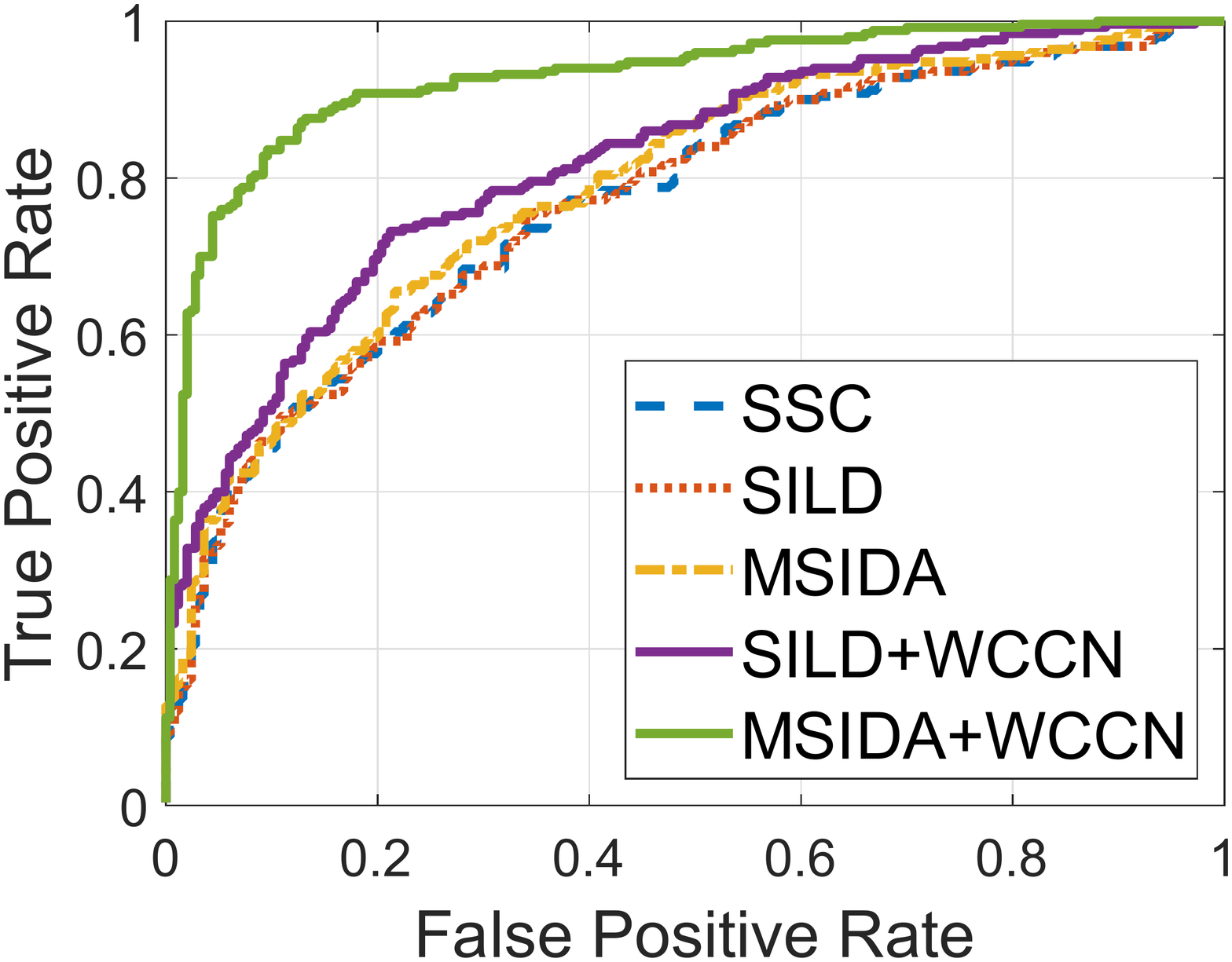}}\quad
   \subfloat[][]{\includegraphics[width=.23\textwidth]{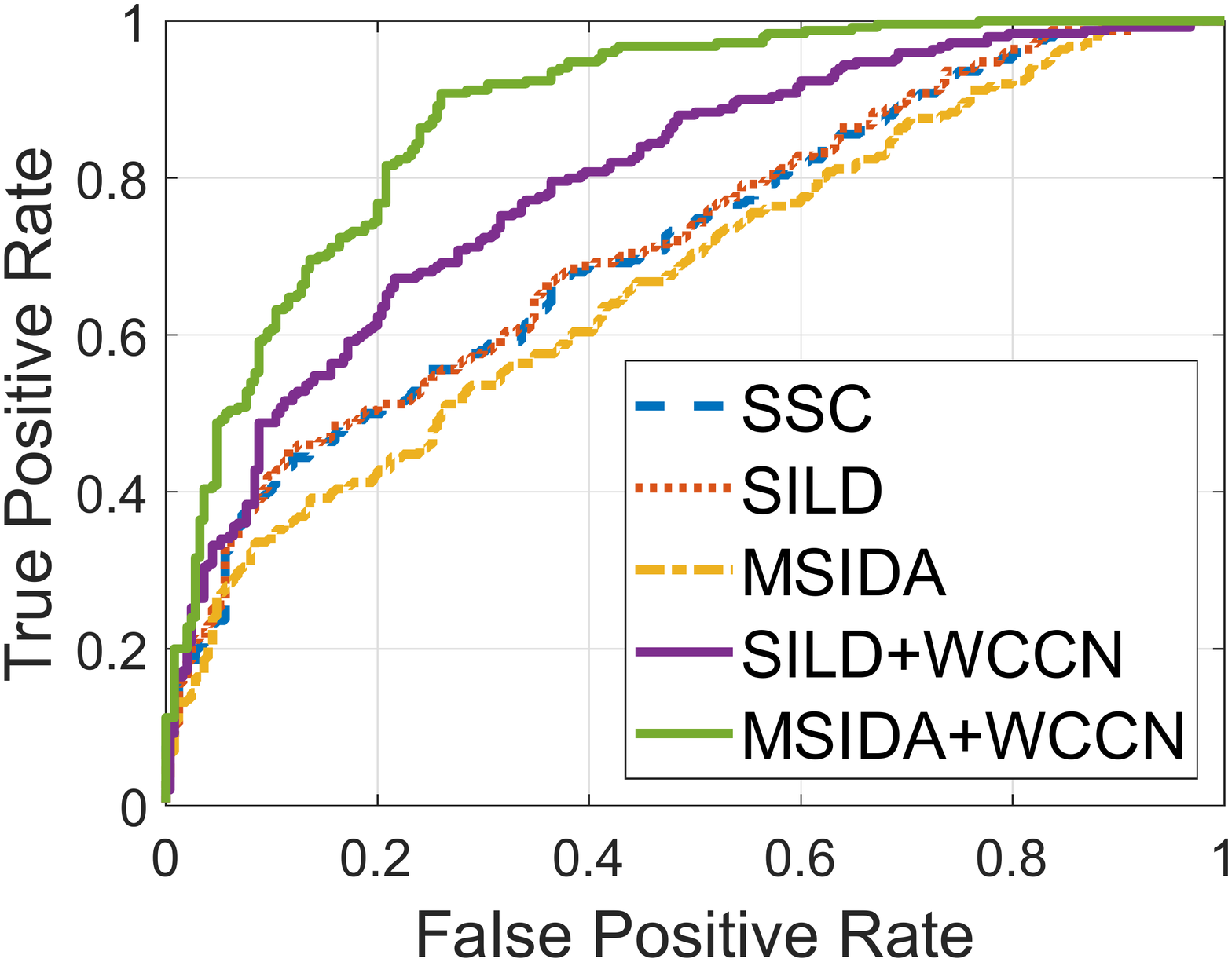}}\quad
   \subfloat[][]{\includegraphics[width=.23\textwidth]{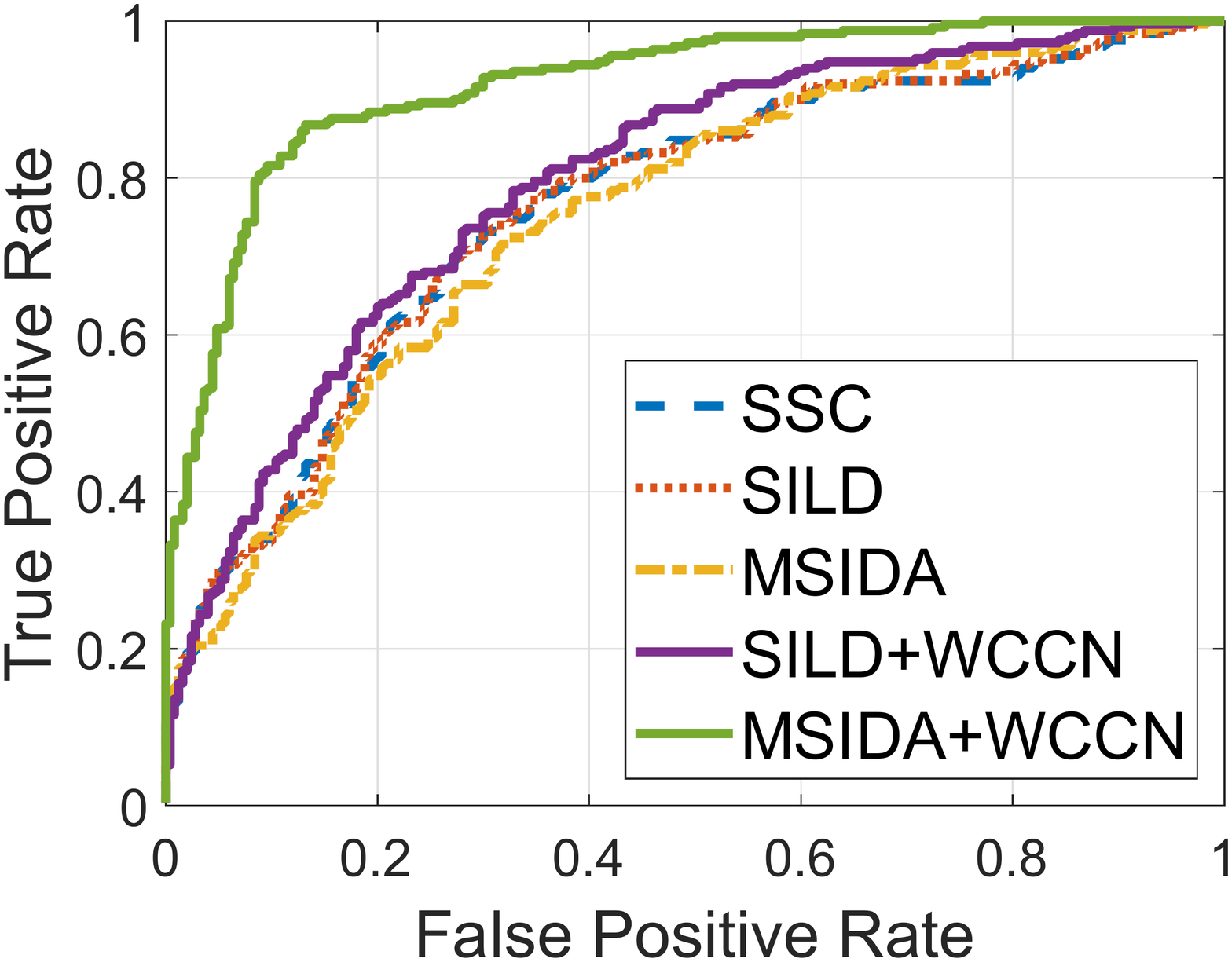}}\quad
   \subfloat[][]{\includegraphics[width=.23\textwidth]{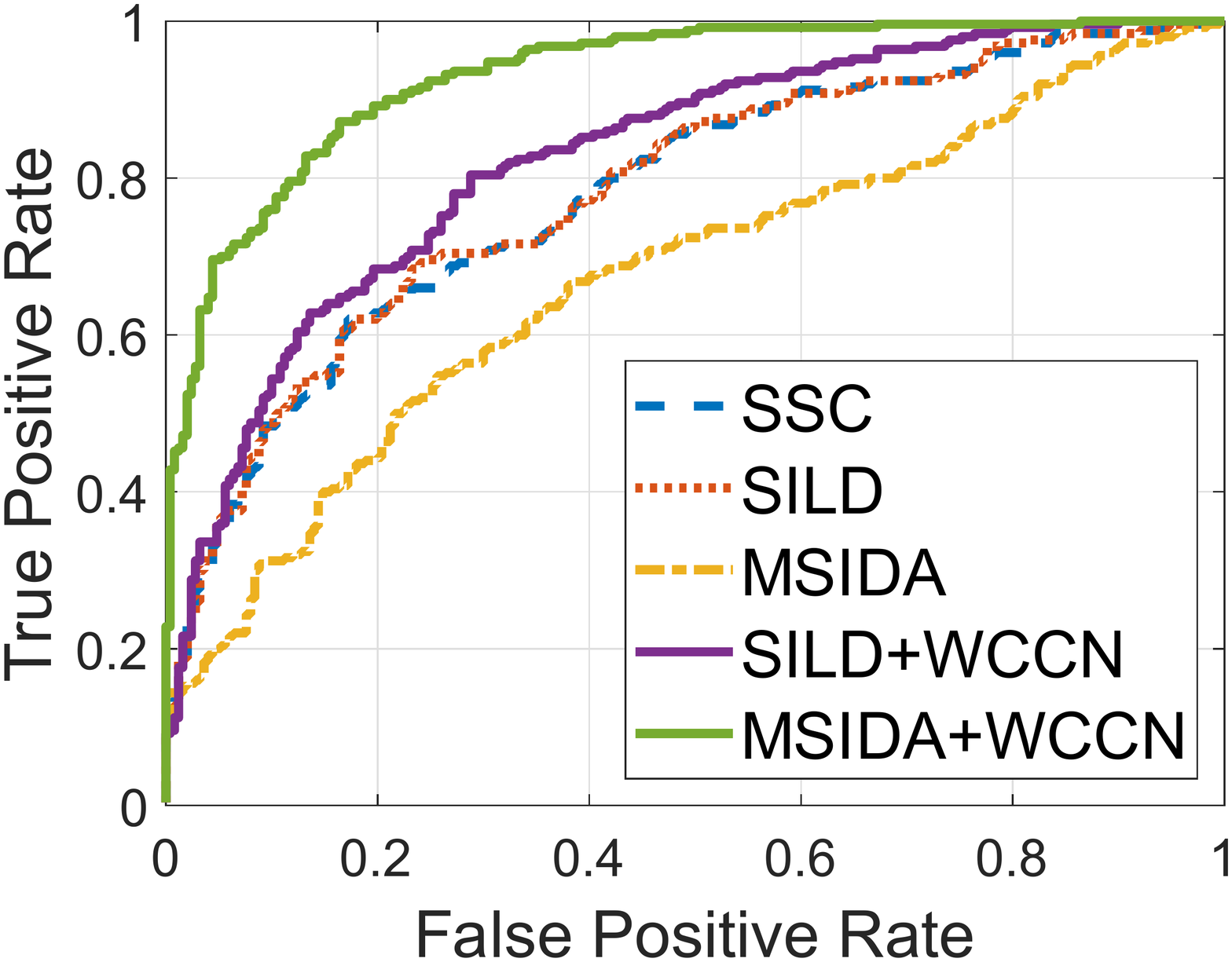}}
   \caption{ROC curves of different methods (SSC, SILD, MSIDA, SILD+WCCN and MSIDA+WCCN) on KinFaceW-II database obtained on (a) F-S set, (b) F-D set, (c) M-S set and (d) M-D set.}
   \label{fig:3}
\end{figure*}

\subsection{Experimental Setup}

The number of the positive and negative pairs used in the
experiments is the same for each relation on the four
subsets. We use five-fold cross validation strategy for the
evaluation. We report the mean accuracy over the five folds.
The negative pairs and folds are predefined for the all four relations.
For the facial deep features and object deep features, we extracted VGG-Face, VGG-F, VGG-M and VGG-S as this has shown to perform better than shallow  methods ~\cite{ZHOU201984,Dornaika2019}. The tensor features are performs by the proposed MSIDA+WCCN method. 

\subsection{Results and Analysis}
\subsubsection{Results on RFIW'20 Challenge}
For RFIW'20 Challenge ~\cite{m1,m2,m3,8337841}, we used the eight fully connected layers of the four pre-trained models (facial and object models). For this reason, we used the Simple Scoring Cosine similarity (SSC) method by concatenating the eight deep features to form a vector of features for each pair facial images. Then, we compute the cosine similarity metric between the two vectors. This method show and prove how can a raw weights of deep features perform in kinship verification as an excellent features of facial images without using any application of learning methods.

\subsubsection{Results on KinFaceW databases}
We run the experiments on the four relations of the two databases, KinFaceW-I and KinFaceW-II, using SSC, SILD, MSIDA, SILD+WCCN and MSIDA+WCCN methods. The results of these experiments are reported in Table \ref{tab:1}. The ROC curves comparing SSC, SILD, MSIDA, SILD+WCCN and MSIDA+WCCN are provided in Figures \ref{fig:2} and \ref{fig:3} for the four relation of KinFaceW-I and KinFaceW-II databases, respectively. As can be noticed from the figure, the performance of MSIDA+WCCN is much better than that the other methods in all cases.

Our proposed method is compared against some recent state-of-the-art methods in Table \ref{tab:1}. Note that some of these methods, such as MvDML, DDMML, ResNet+CF, MNRML+SVM, use combination of different features to describe a face image. Some other methods are based on deep learning. On the four relations of KinFaceW-I and KinFaceW-II databases, our approach yields in the best results for the mean of all the four kinship subsets of the two databases. These results are promising and demonstrate that our proposed approach is performs better than the recent methods for kinship verification. Furthermore, MSIDA+WCCN and SILD+WCCN improve the performances of their counterparts (i.e. MSIDA and SILD) with large margin. Besides, for linear (vector-based) methods, SILD+WCCN improves SILD method with about 2.47\% and 4.30\% for KinFaceW-I and KinFaceW-II databases, respectively. Also, for multilinear (tensor-based) methods, MSIDA+WCCN improves MSIDA method with about 12.08\% and 13.05\% for KinFaceW-I and KinFaceW-II databases, respectively. Thus, the integration of WCCN shows stable and robust performances on the metric learning methods for kinship verification.

\section{Conclusion}\label{conc}
In this paper, we presented an effective approach based on multi-view deep features (facial and object) features to the problem of kinship verification. To achieve a low dimensional and discriminative subspace, we proposed the MSIDA+WCCN method. Also, we studied the effect of WCCN on different metric learning methods showing that the within-class intra-variability introduced by the training data (multi-view deep features in our case) can be reduced to a large extent. Thus, we see that the performances was improved and the metric learning methods can learn good metrics through WCCN integration. The obtained results by MSIDA+WCCN method outperform the state of the art on four Parent-Child relations on two databases, KinFaceW-I and KinFaceW-II.

\bibliographystyle{ieee}
\bibliography{egbib}

\end{document}